\documentclass[11pt]{article}

\usepackage[final]{acl}

\usepackage{times}
\usepackage{latexsym}
\usepackage{amssymb}
\usepackage{amsmath}
\usepackage{subcaption}
\usepackage{booktabs}
\usepackage{multirow}
\usepackage{pifont}
\usepackage{array}
\usepackage[T1]{fontenc}
\usepackage[utf8]{inputenc}

\usepackage{microtype}
\usepackage{xurl}
\usepackage{hyperref}

\usepackage{inconsolata}

\usepackage{graphicx}

\usepackage{tabularx}
\usepackage[table]{xcolor}

\definecolor{CornellRed}{rgb}{0.7, 0.11, 0.11}
\definecolor{OliveGreen}{rgb}{0,0.6,0}

\newcommand{\cmark}{\ding{51}}%
\newcommand{\xmark}{\ding{55}}%

\title{Who Am I? History-Aware Profiles for Student Simulation in Tutoring Dialogues} 

\author{
  \textbf{Zhangqi Duan}\textsuperscript{1},
  \textbf{Shuyan Huang}\textsuperscript{1},
  \textbf{Alexander Scarlatos}\textsuperscript{1},
  \textbf{Jaewook Lee}\textsuperscript{1},\\
  \textbf{Simon Woodhead}\textsuperscript{2},
  \textbf{Andrew Lan}\textsuperscript{1} \\
  \textsuperscript{1}University of Massachusetts Amherst,
  \textsuperscript{2}Eedi \\
  \texttt{\{zduan, shuang, ajscarlatos, jaewooklee, andrewlan\}@cs.umass.edu},\\
  \texttt{simon.woodhead@eedi.co.uk}
}

\begin{document}
\maketitle
\begin{abstract}
A key part of developing large language model (LLM)-powered, automated tutoring tools is student simulation, i.e., using LLMs to role-play as students, which can facilitate tutor model evaluation and training. Existing work mostly focuses on within-dialogue simulation, which lacks context on student knowledge and behavior, partly due to not grounding in past student question-answering or dialogue interactions. In this work, we introduce the task of history-conditioned student simulation, where the goal is to accurately predict student dialogue turns by leveraging information in the student's learning history. We propose a two-component framework in which a profile generator summarizes a student’s history and a simulator predicts student turns conditioned on the resulting profile. We train both components with reinforcement learning (RL), yielding profiles optimized for faithful student simulation. We evaluate our method and baselines on the first-of-its-kind real-world dataset of student dialogues and question responses that we collect from a math learning platform. Extensive experiments show that our method significantly outperforms baselines, and demonstrate the importance of history, profiles, and RL training. 
\end{abstract}

\section{Introduction}

\begin{figure}[t!]
    \centering
    \includegraphics[width=1\linewidth]{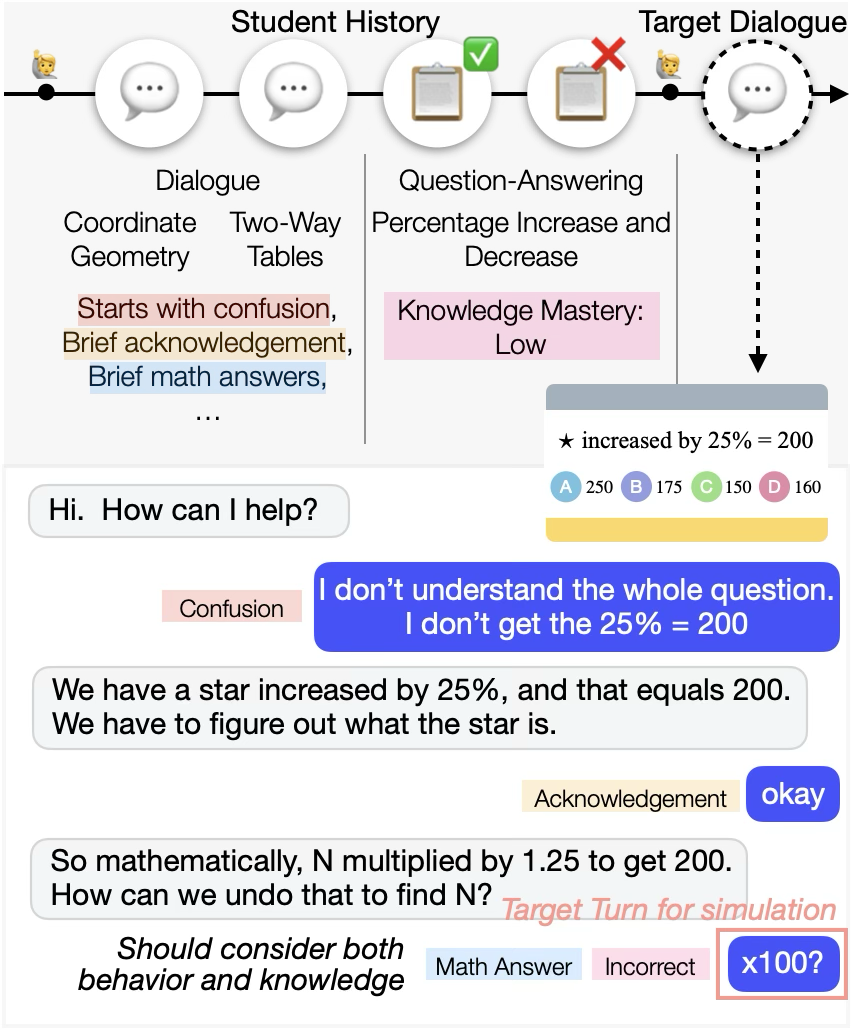}
    \caption{Illustrating the history-conditioned student simulation task. Simulating a student’s next dialogue turn requires information on both knowledge state and behavioral tendencies from their learning history.}
    \label{fig:motivation}
    \vspace{-.5cm}
\end{figure}

Individualized tutoring is a highly effective way to support student learning, motivating recent efforts to scale math tutoring through natural-language dialogue with large language model (LLM)-based tutoring systems \cite{openai-outcomes, team2025ai, wang2024tutor}. In these systems, one important factor for effective adaptation is \emph{student modeling}: estimating a student's evolving knowledge state, including mastery of relevant knowledge components (KCs) and misconceptions, from their dialogue and learning history \cite{wang-etal-2025-training, scarlatos2024exploringknowledgetracingtutorstudent}. Such models can support human tutors through dashboards that surface students' progress and needs \cite{Aleven2016DevelopingAT}, and can also personalize LLM tutors with student-specific context in the prompt. For example, Khan Academy reports that providing Khanmigo with structured signals from a student's learning history, including recent performance and skill gaps, improved students' next-question response correctness by 6.1\% \cite{khanAcademyReport}. 

Complementary to student modeling, \emph{student simulation} aims to reproduce how students behave in tutoring dialogues by generating plausible student turns. Simulated students can provide low-stakes practice environments for human tutor training \cite{cao2026developingauthenticsimulatedlearners} and scalable interaction partners for training and evaluating AI tutors, including through reinforcement learning \cite{eth-tutor-rl,litype}. Prior work on LLM-based student simulation can be classified into two main approaches: prompting and fine-tuning. Prompting-based methods typically first create a synthetic student profile and then prompt a pre-trained LLM to act as that student. The LLM is instructed to exhibit certain misconceptions, personality, or behavioral traits~\cite{liu-etal-2024-personality,lu2024generative,wu2025embracingimperfectionsimulatingstudents}. However, because these simulated students are not based on real students, their responses may differ from what actual students would say. 
%Moreover, these methods often fail to account for individual student histories and lack longitudinal information, making it difficult to capture how a student’s knowledge evolves over time. 
Fine-tuning-based methods, on the other hand, use real tutoring dialogues to train LLMs to generate student utterances~\cite{perczel2025teachlm,scarlatos2026simulatedstudentstutoringdialogues,cao2026developingauthenticsimulatedlearners}. While these approaches are promising because they are grounded in authentic student interactions, their effectiveness remains limited since large-scale dialogue record data remains scarce, making it difficult to capture specifics about each individual student. We provide a more detailed discussion of related work in Appendix~\ref{sec:related-work}.

The key to authentic student simulation is to faithfully reflect a student's \emph{knowledge levels} and \emph{behavioral tendencies} \cite{eth-sim-student}. To capture these, we first need to understand how students interact within learning platforms. As shown in Figure~\ref{fig:motivation}, students may either engage in standalone question-answering practice or seek help from tutors through dialogue. Question-answering records offer explicit evidence of student knowledge, where longitudinal histories indicate how the student’s understanding evolves over time. In addition, dialogue interactions reveal student behavior: how a student responds to tutor questions, whether they tend to give brief answers, and whether they use informal language or emojis. Therefore, accounting for these (possibly long) histories is important to student turn simulation in dialogues.

\paragraph{Contributions} 
In this paper, we establish the new task of \textbf{history-conditioned student dialogue simulation}, i.e., student modeling from learning history and using this information for student simulation in future tutoring dialogues. We identify two key challenges: (1) how to extract informative signals from a student’s prior interactions, including both knowledge state and dialogue behavior; and (2) how to incorporate them into a simulation model to predict the student’s utterance in future dialogue turns. 
First, we build a first-of-its-kind dataset from a popular math learning platform that contains longitudinal student histories, which contain both question-answering records and tutoring dialogues. 
%We focus on this dataset because longitudinal student interaction logs with both dialogue and problem-solving histories are rarely available at the scale and granularity required for our setting. 
%Moreover, such data constitute sensitive educational records and may contain personally identifiable information, making public release and cross-dataset replication difficult.
%
% We conduct extensive experiments comparing our approach with prompting and fine-tuning-based student simulation baselines. Results show that our framework significantly outperforms existing approaches, and ablation studies demonstrate the importance of the question-answering history, dialogue history, profile generation, and our GRPO and DPO training. We additionally discuss a potential dataset as a promising candidate for future evaluation in 
%
%First, we formulate the task of student simulation conditioned on a student's historical interactions. The goal is to extract a profile from a student's history and use it to faithfully simulate that student. 
Second, we propose an end-to-end framework that decomposes student simulation into two coupled components: a \textbf{profile generator} and a \textbf{student simulator}\footnote{Our code and data can be found at: \url{https://github.com/umass-ml4ed/dialogue_student_sim}.}. The profile generator summarizes the relevant part of a student’s learning history into a structured profile, capturing both knowledge state and behavioral tendencies. The student simulator then conditions on this profile to generate future dialogue turns that are consistent with the student’s expected performance and behavior. Third, we use RL to optimize both the generated profiles and the simulator to make downstream simulations more faithful. Across ablations and case studies, we report several findings, including the key one: question-answering and dialogue histories are two complementary views of the same student and are both key to faithful student simulation. Question-answering history not only reveals knowledge levels and misconceptions, but also affects how students respond in dialogue; dialogue history not only reveals student behavior, but also adds detailed, fine-grained insights into knowledge. %Thus, faithful student simulation requires profiles that jointly encode what a student knows and how that student behaves.

\begin{figure*}
    \centering
    \includegraphics[width=1\linewidth]{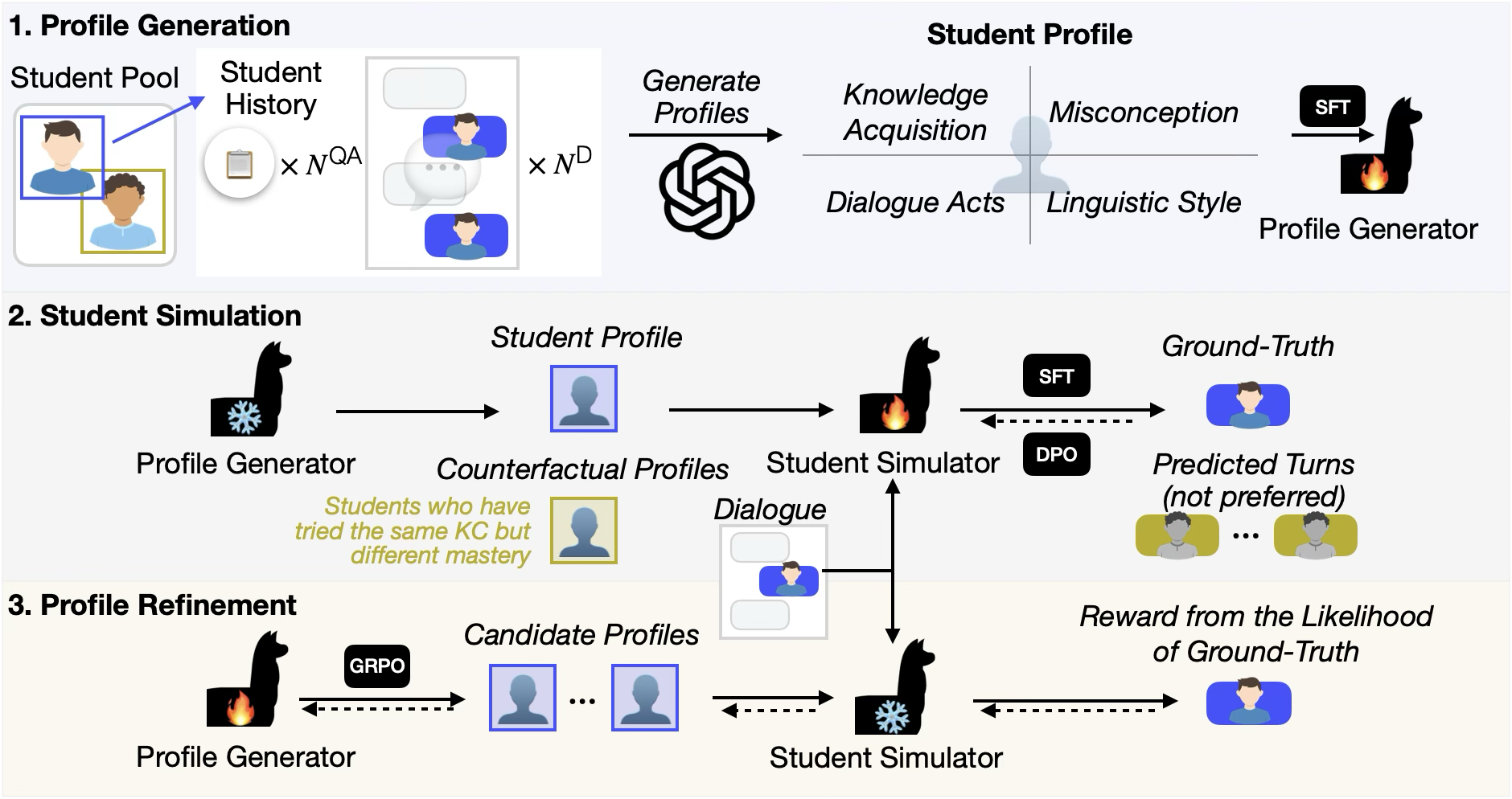}
    \vspace{-.5cm}
    \caption{An overview of our framework. First, we summarize a student's relevant question-answering and dialogue interactions into a profile. Second, we use this profile in the student simulator to simulate student turns in future dialogues. We refine both the profile generator and student simulator by optimizing them via RL.}
    \label{fig:method-overview}
    \vspace{-.5cm}
\end{figure*}

\section{Task Definition \& Data}
We now formally define the task of history-conditioned student simulation in math tutoring dialogues. We consider a learning platform in which students attempt math questions. When a student is stuck or answers a question incorrectly, they can initiate a dialogue with a tutor grounded in the question. The tutor then guides the student through the mathematical reasoning process to solve the question. Our goal is to faithfully simulate the student turns in the dialogue context, conditioned on their past learning history that contains both question-answering and dialogue records.

For a student $u$, we represent the ordered learning history as
$H_u = (x_{u,1}, \ldots, x_{u,T}).$
Each interaction $x_{u,j}$ is either question-answering, defined as
$x_{u,j}^{\mathrm{QA}}=(q_{u,j}, a_{u,j})$,
where $q_{u,j}$ is a multiple-choice math question and $a_{u,j} \in \{0,1\}$ indicates whether the student's answer is correct, or dialogue, defined as
$x_{u,j}^{\mathrm{D}}=(q_{u,j}, d_{u,j})$,
where $d_{u,j}$ is a tutoring dialogue grounded in question $q_{u,j}$. We represent each dialogue as an alternating sequence of student and tutor turns,
$d = (s_0, t_1, s_1, \ldots, t_M, s_M)$,
where $s_i$ is the student utterance at turn $i$, $t_i$ is the tutor utterance, and $M$ is the number of tutor-student turn pairs. The initial turn $s_0$ is present when the student initiates the conversation, and the final turn $s_M$ is present when the student ends the dialogue. Each question $q$ is associated with a set of KCs, $C(q)$, corresponding to the skills needed to answer the question correctly. This interaction-based notation reflects the natural structure of tutoring platforms: students may solve some questions independently but request tutoring support for others.

For any target timestep $j$, we denote the student's prior history as $h_{<j}^u=(x_{u,1},\ldots,x_{u,j-1})$. Let $\mathcal{M}_S$ denote the student simulator. At each turn $i$, we generate the next student utterance
$\hat{s}_i \sim \mathcal{M}_S(s \mid s_{<i}, t_{\leq i}, q_{u,j}, h_{<j}^u)$
and compare it against the ground-truth response $s_i$. In practice, directly conditioning on $h_{<j}^u$ is often infeasible, since student histories can be long and contain interactions that are irrelevant to the current question. We may therefore need a profile generator, which compresses $h_{<j}^u$ into a compact textual profile $p_{u,j}$, and the simulator generates
$\hat{s}_i \sim \mathcal{M}_S(s \mid s_{<i}, t_{\leq i}, q_{u,j}, p_{u,j})$.
The key challenge is to produce profiles that preserve both the knowledge state and behavioral tendencies needed for faithful turn-level simulation in future dialogues.

\section{Methodology}
We now detail our methodology for history-conditioned student simulation in future dialogues. As shown in Figure~\ref{fig:method-overview}, our framework has three stages. First, we generate textual student profiles based on each student's prior learning history. Second, we train a profile-conditioned student simulator to predict student utterances, using supervised fine-tuning (SFT) followed by preference optimization. Finally, we refine the profile generator with RL, using downstream dialogue simulation performance as the reward signal. Our framework is conceptually related to the method in \cite{li-etal-2025-cikt}: both leverage a profile generator optimized on downstream task performance. However, that work focuses on knowledge tracing, i.e., next-question correctness prediction given question-answering history. We focus on dialogue simulation and history that contains both dialogues and question-answering records, leading to completely different profile structures. We discuss the detailed comparison in Appendix~\ref{sec:related-work}.

\subsection{Student Profile Generator}
We first detail how we construct student profiles, which serve as compact representations of learning histories for student simulation.

\paragraph{Initial Profile Generation} 
For each student $u$ at timestep $j$, we construct a profile $p_{u,j}$ conditioned on the prior history $h_{<j}^u$. The profile is intended to be a compact, interpretable summary of the student's current knowledge and dialogue behavior.
% , since including the whole history records is implausible.
We initially construct profiles by prompting a proprietary reasoning LLM given the student's learning history. Practically, a student may have long learning interactions. Due to the context length limitations of LLMs, and motivated by prior work suggesting that not all historical interactions are equally predictive, we use a compact subset of recent interactions for profile generation~\cite{huang2023towards, li2024explainable}. Specifically, we select the most recent $N^{\mathrm{QA}}$ question-answering interactions, including associated KCs and the student's correctness, and $N^{\mathrm{D}}$ dialogue interactions before timestep $j$ to summarize student's past learning behavior. In this paper, we set $N^{\mathrm{QA}}=8$ and $N^{\mathrm{D}}=3$ as a trade-off between context-length limitations and retaining sufficient historical information. We provide analysis of this setting in Appendix~\ref{app:history-window}.

Each generated profile has a fixed structure with five components, which align with important aspects of student knowledge and behavior identified in~\cite{scarlatos2026simulatedstudentstutoringdialogues}. First, a \textit{Knowledge State} section reports estimated mastery for all KCs, computed as correct attempts over total attempts across mapped questions. Second, a \textit{Knowledge Acquisition} section describes how the student's mastery appears to change over time, highlighting improvement, stagnation, or decline across attempts and dialogues. Third, a \textit{Misconception} section records recurring mathematical errors, such as arithmetic slips, equation setup errors, or misunderstandings of particular concepts. Fourth, a \textit{Dialogue Behavior} section describes the student's tendencies and common dialogue acts in tutoring dialogues. Fifth, a \textit{Linguistic Style} section captures surface-level interaction patterns, such as verbosity, use of mathematical terminology, and whether the student tends to answer with full explanations or short fragments. These prompted profiles serve as an initialization point for the trainable profile generator described in Section~\ref{sec:profile-generator-training}. We show the prompt used for profile generation in Appendix~\ref{sec:prompts}.
% across the five act types: math answering, not understanding, seeking information, off-topic responses, and acknowledgment

\subsection{Student Simulator} 
\label{sec:student-simulator}
We use the student dialogue simulator $\mathcal{M}_S$ to predict a student's next utterance at each turn in a tutoring dialogue. Given the grounded question $q_{u,j}$, the generated student profile $p_{u,j}$, and the dialogue context up to turn $i$, the model generates a simulated student utterance $\hat{s}_i$. Formally,
$\hat{s}_i \sim \mathcal{M}_S(\,s \mid s_{<i}, t_{\leq i}, q_{u,j}, p_{u,j})$.
We initialize $\mathcal{M}_S$ by performing SFT on observed, ground-truth tutor-student dialogues. 

\paragraph{Profile-Aligned Preference Optimization}
In order to improve the alignment of the simulated student model with the given profiles, we perform an additional training phase leveraging preference optimization. We construct preference pairs by contrasting each observed student turn with counterfactual alternatives generated under mismatched profiles. This formulation encourages the model to distinguish not only between generally plausible responses, but also between responses that are inconsistent with the intended student profile.

For each student turn $s_i$ in a dialogue grounded in $q_{u,j}$ and associated with profile $p_{u,j}^{+}$, we retrieve all valid counterfactual profiles $P_{u,j}^{-}$ that contain at least one KC relevant to $q_{u,j}$ and differ from $p_{u,j}^{+}$ by at least a threshold $\delta$ in terms of mastery level on that KC. Formally, $P_{u,j}^{-} = \{ p_{u,j}^{-} \mid \exists c \in C(q_{u,j}) \cap C(p_{u,j}^{+}) \cap C(p_{u,j}^{-}) \quad \text{s.t.} \quad |z_{p_{u,j}^{+}}(c)-z_{p_{u,j}^{-}}(c)| \geq \delta \}$,
where $C(p)$ is the set of KCs with knowledge state estimates in $p$,
$z_p(c)=\frac{n_p^{\mathrm{correct}}(c)}{n_p^{\mathrm{attempt}}(c)}$,
and $n_p^{\mathrm{correct}}(c)$ and $n_p^{\mathrm{attempt}}(c)$ are the number of correct and total attempts for $c$ in $p$, respectively.
We then generate counterfactual responses by conditioning the SFT simulator on each $p_{u,j}^{-} \in P_{u,j}^{-}$ while keeping the same dialogue context, and sample $K$ responses for turn $i$: $\tilde{s}_{i}^{(k)} \sim \mathcal{M}_S(s \mid s_{<i}, t_{\leq i}, q_{u,j}, p_{u,j}^{-})$ for $k \in \{1,\ldots,K\}$.

We then score each counterfactual response on our evaluation metrics (which we detail later in Section~\ref{sec:metrics}), to form a reward $r(\tilde{s}^{(k)}_i)$, excluding the linguistic metrics since they can be highly inconsistent depending on the dialogue context. If $r(\tilde{s}^{(k)}_i)<\tau$ for a preset threshold $\tau$, we classify $\tilde{s}^{(k)}_i$ as sufficiently different from the ground-truth turn and form a preference pair $(x_i,s_i,\tilde{s}^{(k)}_i)$, where $x_i=(s_{<i},t_{\leq i},q_{u,j},p_{u,j}^{+})$ is the original context. This way, the preference pairs are highly relevant yet not too noisy. We train $\mathcal{M}_S$ using direct preference optimization (DPO) \cite{rafailov2023direct}, where $s_i$ is preferred and $\tilde{s}^{(k)}_i$ is dispreferred. As a result, this procedure trains the simulator to generate turns that are aligned with the given profile, using faithfulness in the downstream simulation task as the alignment criterion. 

\subsection{Profile Generator Tuning via RL}
\label{sec:profile-generator-training}
The final stage trains the profile generator. We first distill the prompted profiles into a trainable open-source LLM $\mathcal{M}_P$, making it acquire the profile structure we designed without repeatedly prompting a proprietary LLM. Specifically, we use historical context $h_{<j}^u$ as input and perform SFT on profiles generated in the initialization stage, $p_{u,j}$. 

We further refine $\mathcal{M}_P$ with group relative policy optimization (GRPO) \cite{grpo}, to optimize for faithful downstream dialogue simulation. 
%Here, we simply use the likelihood of the ground-truth student dialogue turns under the simulator model as the reward. 
Our goal is to encourage the generator to produce profiles that are not only informative summaries of past history, but also useful representations to reproduce the student's future dialogues. 
Given a history context $h_{<j}^u$ for student $u$ at a timestep $j$ that is a dialogue interaction, the current profile generator samples a group of $G$ candidate profiles
$\{p_{u,j}^{(1)}, \ldots, p_{u,j}^{(G)}\}$.
Each candidate profile is used as conditioning context for the fixed student dialogue simulator $\mathcal{M}_S$. We define the dialogue reward as the average log-likelihood of the ground-truth student turns:
$r_{u,j,\mathrm{D}}^{(g)} = \frac{1}{M+1} \sum_{i=0}^{M} \log \mathcal{M}_S\big(s_i \mid s_{<i}, t_{\leq i}, q_{u,j}, p_{u,j}^{(g)}\big)$.
%This reward measures whether the profile contains information that helps reproduce the student's actual dialogue behavior. 
GRPO training then updates $\mathcal{M}_P$ to increase the probability of profiles with positive relative advantage and decrease the probability of profiles with negative relative advantage, while constraining the update to stay close to the reference model.
After GRPO training, we use the updated profile generator to produce refined profiles for each student history and evaluate on the downstream dialogue simulation task.  

\section{Experimental Setup}

We now detail our experimental setup to validate the effectiveness of our framework. 

\subsection{Dataset}

We conduct experiments on a dataset we build from a real-world online learning platform, where students practice math multiple-choice questions and can initiate chat-based conversations with a human tutor in an on-demand way, at any point during question-answering interactions. Questions span a wide range of topics, including Algebra, Geometry, and Number. After data processing, the dataset contains $1,775$ dialogues and $66,705$ question-answering records from 670 students. Additional details are available in the Appendix~\ref{apdx:ds_details}; we are in the process of publicly releasing the dataset after careful PII removal from the dialogues.

\begin{table*}[t]
\centering
\small
\setlength{\tabcolsep}{9pt}
\renewcommand{\arraystretch}{1.12}
\begin{tabular}{lccccc}
\toprule
Method & Acts & Corr. & Errors & Cos. Sim. & ROUGE-L \\
\midrule
\multicolumn{6}{c}{\cellcolor{gray!20}Simulator Base Model: \textbf{GPT-5.4}} \\
\midrule
Knowledge Profile & $0.427_{\pm 0.010}$ & $0.438_{\pm 0.071}$ & $0.086_{\pm 0.012}$ & $0.446_{\pm 0.035}$ & $0.104_{\pm 0.019}$ \\

OCEAN Persona& $0.443_{\pm 0.016}$ & $0.456_{\pm 0.038}$ & $0.079_{\pm 0.027}$ & $0.465_{\pm 0.027}$ & $0.120_{\pm 0.014}$ \\

Cognitive Profile & $0.475_{\pm 0.009}$ & $0.461_{\pm 0.018}$ & $0.105_{\pm 0.009}$ & $0.503_{\pm 0.033}$ & $0.150_{\pm 0.013}$ \\

History ICL & $0.495_{\pm 0.036}$ & $0.466_{\pm 0.031}$ & $0.094_{\pm 0.022}$ & $0.531_{\pm 0.041}$ & $0.176_{\pm 0.036}$ \\

\midrule
\multicolumn{6}{c}{\cellcolor{gray!20}Simulator Base Model: \textbf{Llama-3.1-8B-Instruct}} \\
\midrule
SFT & $0.586_{\pm 0.021}$ & $0.465_{\pm 0.032}$ & $0.102_{\pm 0.017}$ & $0.680_{\pm 0.018}$ & $0.261_{\pm 0.034}$ \\

DPO  & $0.593_{\pm 0.019}$ & $0.474_{\pm 0.029}$ & $0.108_{\pm 0.015}$ & $0.680_{\pm 0.022}$ & $0.259_{\pm 0.031}$ \\

History SFT  & $\underline{0.602}_{\pm 0.014}$ & $\underline{0.479}_{\pm 0.022}$ & $\underline{0.111}_{\pm 0.021}$ & $\underline{0.694}_{\pm 0.030}$ & $\underline{0.287}_{\pm 0.026}$ \\

ProfileRL (Ours) & $\textbf{0.644}_{\pm 0.023}^{*}$ & $\textbf{0.516}_{\pm 0.027}^{*}$ & $\textbf{0.149}_{\pm 0.018}^{*}$ & $\textbf{0.700}_{\pm 0.017}$ & $\textbf{0.296}_{\pm 0.023}$ \\
\bottomrule
\end{tabular}
\caption{Results for history-conditioned student simulation. Best method is \textbf{bolded} and second best is \underline{underlined}. $^{*}$ denotes statistically significant improvement over all baselines ($p<0.05$).}
\label{tab:dialogue-metrics}
\end{table*}

\subsection{Baselines}

We compare our framework, which we name \textbf{ProfileRL} for convenience, to strong prompting and fine-tuning baselines, to examine what aspects of student profiles are beneficial to the downstream student dialogue simulation task.

For prompting baselines, most existing works construct synthetic personas to induce behavioral diversity rather than grounding the persona in a real student's learning history. We adapt these methods to our task and provide learning histories in different formats to the LLM. These baselines include:

\noindent\textbf{Knowledge Profile} We adapt the knowledge state profile of \citet{lu2024generative}, which summarizes a student's mastery on KCs. Specifically, we estimate each observed KC as ``mastered'', ``confused'', or ``unknown'' based on the student's prior question-answering interactions, and provide the resulting profile to the simulation prompt as a compact representation of the student's knowledge. 

\noindent\textbf{OCEAN Persona} We adapt the persona-based approach of \citet{liu-etal-2024-personality}, which represents a simulated student using Big Five personality traits and language ability. For each student, we first prompt an LLM to infer the five OCEAN traits and language proficiency as either ``low'' or ``high'' from the student's dialogue history. We then provide this estimated persona to the student-simulation prompt and instruct the LLM to follow the traits.

\noindent\textbf{Cognitive Profile} We adapt the cognitive-profile prompting method of \citet{wu2025embracingimperfectionsimulatingstudents}. This baseline constructs a profile that describes both the student's mastery of individual KCs and the relationships among those KCs. Following prior work, we first prompt an LLM to identify KCs at multiple granularities from the questions and to build a relation graph over KCs across questions. We then prompt the LLM to estimate the student's mastery of each KC based on their history. During simulation, the model first predicts the student's next dialogue act as a chain-of-thought reasoning step and then generates the dialogue turn utterance.

\noindent\textbf{In-Context Learning (ICL)} We provide the full prior dialogue and question-answering history from the target student as in-context examples and prompt the model to simulate student utterances based on the patterns shown in the examples.

In addition, we also compare against several fine-tuning baselines. \textbf{SFT} trains a student simulator with SFT on observed student turns \citep{perczel2025teachlm}. We also compare with \textbf{History SFT}, which fine-tunes an LLM given both the question-answering and dialogue interactions in each student's learning history, using the same truncated history window as the profile generator.  Moreover, we compare with \textbf{DPO} \cite{scarlatos2026simulatedstudentstutoringdialogues}, which first overgenerates candidate student turns and scores them based on evaluation metrics, and then constructs preference pairs based on high quality turns to train the model with DPO.

\subsection{Experimental Setup}
We report the average performance on $5$-fold cross-validation. We use a 80/10/10 train/validation/test split across \emph{students} to evaluate generalization to students not seen at train-time. 
For fair comparison, we use the same base LLM, Llama-3.1-8B-Instruct~\cite{grattafiori2024llama3herdmodels}, for fine-tuned baselines, and evaluate prompting-only baselines using GPT-5.4~\cite{openai2026gpt54}. We train all models using LoRA \cite{hu2021loralowrankadaptationlarge}. 
We provide full experimental details in Appendix~\ref{apdx:exp_details}.

\subsection{Metrics}
\label{sec:metrics}
We follow the simulated student evaluation protocol from \citep{scarlatos2026simulatedstudentstutoringdialogues}, where simulated turns are compared to ground-truth turns across multiple reference-based metrics. We focus on five such turn-level metrics that measure similarity among dialogue acts, response correctness, misconceptions, and linguistic similarity. 

\textbf{Acts} measures whether the simulated turn expresses the same dialogue
act as the ground-truth turn. The full dialogue act label set is given in Appendix~\ref{apdx:ds_details}. \textbf{Correctness} (Corr.) measures whether the
simulated and ground-truth turns have the same correctness status with respect to the preceding tutor turn, i.e., whether both are correct or both are
incorrect. \textbf{Errors} is computed for ground-truth turns that are incorrect and measures whether the simulated turn reflects the same underlying
mistake or misconception. Each of these three metrics is binary at the turn level.

We also report two text-based similarity metrics between simulated and ground-truth turns. \textbf{Cosine Similarity} (Cos.\ Sim.) measures semantic
similarity between their text embeddings, while \textbf{ROUGE-L} measures word-level recall based on the longest common subsequence \citep{lin-2004-rouge}.

\begin{table*}[t]
\centering
\small
\resizebox{\textwidth}{!}{
\begin{tabular}{lccc ccc ccccc}
\toprule
\multirow{2}{*}{Method}
& \multicolumn{3}{c}{Input Information}
& \multicolumn{3}{c}{Training Strategy}
& \multicolumn{5}{c}{Performance} \\
\cmidrule(lr){2-4} \cmidrule(lr){5-7} \cmidrule(lr){8-12}
& Know. & Behav. & Raw Hist.
& SFT & DPO & GRPO
& Acts & Corr. & Errors & Cos. Sim. & ROUGE-L \\
\midrule
ProfileRL (Ours)
& \cmark & \cmark & \xmark
& \cmark & \cmark & \cmark
&  \textbf{0.605} & \textbf{0.466} & \textbf{0.150} & \textbf{0.689} & \textbf{0.271} \\

\quad w/o Simulator DPO
& \cmark & \cmark & \xmark
& \cmark & \xmark & \cmark
& 0.602 & 0.461 & 0.142 & 0.686 & 0.270 \\

\quad w/o Profile GRPO
& \cmark & \cmark & \xmark
& \cmark & \cmark & \xmark
& 0.593 & 0.457 & 0.143 & 0.687 & 0.268 \\

\quad w/o RL Alignment
& \cmark & \cmark & \xmark
& \cmark & \xmark & \xmark
& 0.589 & 0.436 & 0.115 & 0.682 & 0.264 \\

\quad w/o Profile Generator
& \xmark & \xmark & \cmark
& \cmark & \xmark & \xmark
& 0.575 & 0.428 & 0.112 & 0.680 & 0.266 \\

\quad Prompting Only
& \cmark & \cmark & \xmark
& \xmark & \xmark & \xmark
& 0.502 & 0.446 & 0.109 & 0.488 & 0.145 \\
\midrule
\quad w/o Behavior Profile
& \cmark & \xmark & \xmark
& \cmark & \cmark & \cmark
& 0.578 & 0.454 & 0.142 & 0.677 & 0.251 \\

\quad w/o Knowledge Profile
& \xmark & \cmark & \xmark
& \cmark & \cmark & \cmark
& 0.598 & 0.448 & 0.135 & 0.682 & 0.270 \\
\bottomrule
\end{tabular}
}
\caption{
Ablation study of ProfileRL. 
Know., Behav., and Raw Hist.\ indicate the use of summarized knowledge-related profile components, summarized behavior-related profile components, and raw question-answering/dialogue history, respectively. 
GRPO and DPO correspond to training profile generator and student simulator via RL, respectively. 
RL Alignment refers to using both together. w/o Profile Generator is equivalent to History SFT.
}
\label{tab:ablation}
\end{table*}

\section{Results}
We now detail our experimental results, including quantitative results, an ablation study, and a qualitative analysis of generated student profiles and how they help predict student utterances.

\subsection{Quantitative Results}
Table~\ref{tab:dialogue-metrics} shows the average performance and standard deviation for turn-level student simulation. We compare prompting-based baselines, fine-tuned methods without history, and history- or profile-conditioned methods. Overall, methods with student-specific historical information perform best: History SFT improves over SFT and DPO across all metrics, while ProfileRL achieves the strongest overall performance.

Prompting-based methods perform worse than fine-tuning-based models overall, even when conditioned on student information. Correctness is the only metric where they are close: History ICL reaches 0.466, close to SFT at 0.465 and DPO at 0.474. However, this result is partly due to correct responses being the majority class in the ground-truth turns, as shown in Appendix~\ref{apdx:ds_details}; pretrained models tend to generate correct answers, making this metric less indicative of student simulation quality. The much lower performance on the other metrics shows that prompting alone struggles to reproduce students’ dialogue behavior, misconception patterns, and language style, even after taking historical information into account. Methods using only a narrow type of student trait are also insufficient: OCEAN Persona captures personality-related tendencies but not knowledge state, while the Knowledge Profile captures knowledge mastery levels but misses behavioral patterns. These results suggest that faithful student simulation requires comprehensive representations of student history.

Our proposed method, ProfileRL, outperforms all baselines on every metric. Compared to History SFT, ProfileRL improves Acts by 0.042, Corr.\ by 0.037, and Errors by 0.038, with statistically significant gains on these three metrics ($p < 0.05$ under a paired $t$-test), which are most directly related to student behavior and knowledge state modeling. The smaller but consistent gains on Cos.\ Sim.\ and ROUGE-L suggest that profiles effectively capture cognitive and behavioral signals, not just lexical overlap with the reference utterance. Compared with full histories, profiles provide compact summaries of students' knowledge states and linguistic tendencies, filtering noisy context into information relevant to the next dialogue turn. To further assess the generalizability of ProfileRL across model scales, we conduct additional experiments using a smaller student simulator, Llama-3.2-3B. ProfileRL consistently outperforms all prompting and fine-tuning baselines, further supporting the robustness of our approach. Detailed results are reported in Appendix~\ref{appdix: smaller_model}.

Most importantly, question-answering and dialogue histories provide complementary signals that transfer across metric categories, rather than only improving the most targeted metrics. Question-answering records reveal students' knowledge states and misconceptions, which naturally helps knowledge-related metrics such as Corr.\ and Errors. However, this information also helps predict dialogue acts, because students’ knowledge levels shape how they behave in tutoring dialogues: for example, students with high prior knowledge tend to seek help or tell the tutor that they do not understand certain concepts less often. Conversely, dialogue histories tell us more than linguistic style: tutor scaffolding in prior dialogues exposes how students respond to subquestions and hints, providing additional, fine-grained information on their knowledge. Thus, question-answering and dialogue history are complementary rather than isolated views of each student. In contrast, the reason why prior student dialogue simulation work found that RL provides only marginal gains over SFT \cite{scarlatos2026simulatedstudentstutoringdialogues} is likely due to not incorporating student learning history. In that case, there is not enough information on both knowledge and behavior for each student from history to condition the simulation of future dialogues on. 

% \ml{we should highlight the ``aha'' part of our findings: QA provides information on knowledge and it helps acts prediction too - the point here is to explicitly compare against the ACL paper, which does not consider history so RL didn't really help} \zd{Findings added}

\subsection{Ablation Study}
We conduct an ablation study on one data split to understand the contribution of each stage in our training pipeline and each dimension of the generated profile. We show results in Table~\ref{tab:ablation}. 

We first ablate the training stages in the pipeline. Removing both DPO for dialogue simulator training and GRPO for profile generator training leads to the largest drop in performance, decreasing Acts from 0.605 to 0.589, Corr. from 0.466 to 0.436, and Errors from 0.150 to 0.115. This result shows that SFT alone is not sufficient for faithful student simulation, especially when the model needs to match how students respond, whether they answer correctly, and what errors they make. Removing only GRPO for profile generator training also hurts performance on most metrics. This result shows that optimizing the profile generator with the student turn likelihood as the reward is key: without this step, the profile generator does not know which part of the student's history is predictive of their future dialogue responses. Removing only DPO leads to smaller but still consistent drops. The poor performance of History SFT, which removes all three components of our method, shows the importance of using a compact, RL-tuned profile. ProfilePrompt, which uses the profile in a prompting-only method, results in the worst performance, which suggests that the profile can only be effectively utilized when fine-tuned on real student data. Overall, these results verify that our two optimization stages are complementary: DPO makes the base dialogue simulator more faithful, while GRPO makes the profile generator effective in extracting relevant information in student history.

We next ablate the content of the generated profile. We divide the profile into two parts, one focusing on knowledge and the other on behavior: (1) knowledge state, knowledge acquisition, and misconceptions, and (2) dialogue acts and linguistic style. Removing the behavior sections primarily hurts behavior-sensitive metrics, decreasing Acts, Cos.\ Sim., and ROUGE-L, but it also lowers Corr.\ and Errors despite retaining the full knowledge profile. This suggests that how students express answers in dialogue provides cues about whether their responses are correct and what misconceptions they reflect. Conversely, removing the knowledge sections causes the largest drops on Corr.\ and Errors, while also hurting Acts and Cos.\ Sim.\ performance. This indicates that students' dialogue acts and wording also depend on whether they know the underlying concept. These cross-metric drops show that knowledge and behavior provide complementary signals rather than isolated information: a student's knowledge state shapes how they respond in dialogue, while their behavioral patterns provide additional information about correctness and misconceptions. Overall, faithful student simulation requires both dimensions of the profile. We further ablate each profile section and the amount of historical context used for profile generation in Appendix~\ref{appdix: add_ablation}.

\subsection{Qualitative Analysis}
\paragraph{GRPO training helps the downstream simulation task} We now use a qualitative case study to illustrate how GRPO training changes the GPT-generated student profile and why these changes better support faithful turn-level student simulation. The contrasting profiles are shown in Table~\ref{tab:profile_case_study}. We see that the GRPO-refined profile provides more task-relevant evidence to directly guide dialogue simulation. First, it contains a concrete misconception tied to the student's past struggles on \textit{Squares, Cubes, etc.} In contrast, the GPT-generated profile states that no recurring misconception is present, thereby omitting an error pattern that is useful for predicting future student behavior. Second, the GRPO-refined profile shows specific linguistic styles of the student using observable markers from past dialogues, such as ``i dont understand'' and ``am i correct,'' which gives the simulator clearer cues for reproducing the student's uncertainty, brevity, and informal phrasing. Third, the GRPO-refined profile explicitly captures the student's dialogue acts tendency: it specifies when \textit{Acknowledge}, \textit{Not Understanding}, and \textit{Off-Topic} appear, and notes the student does not tend to \textit{Seek Information}. The GPT profile instead summarizes the student's behavior at a high level. These changes suggest that RL is effective, with the reward, i.e., average log-likelihood of the ground-truth student turn conditioned on the GRPO-refined profile being $-1.953$ compared to $-2.056$ for the GPT profile. Overall, GRPO training shifts the profile from a broad summary toward a more diagnostic, behaviorally grounded representation, showing qualitatively why faithful simulation requires the profile to connect what the student knows with how the student expresses themselves in dialogue. We provide one additional head-to-head profile comparison in Appendix~\ref{apdx:add_qualitative}.

\begin{table*}[t]
\centering
\small
\setlength{\tabcolsep}{4pt}
\renewcommand{\arraystretch}{1.15}
\begin{tabularx}{\textwidth}{p{0.15\textwidth} X X}
\toprule
\textbf{Profile Section} & \textbf{GPT Profile} & \textbf{GRPO-refined Profile} \\
\midrule
\textbf{Knowledge State}
&
Factors and Highest Common Factor: 2/2; Multiples and Lowest Common Multiple: 1/1; Factors and Highest Common Factor, Multiples and Lowest Common Multiple: 1/1; Prime Numbers and Prime Factors: 1/2; \textcolor{CornellRed}{Squares, Cubes, etc.: 1/3}.
&
Factors and Highest Common Factor: 2/2; Multiples and Lowest Common Multiple: 1/1; Factors and Highest Common Factor, Multiples and Lowest Common Multiple: 1/1; Prime Numbers and Prime Factors: 1/2; \textcolor{OliveGreen}{Squares, Cubes, etc.: 1/3}.
\\

\textbf{Knowledge Acquisition}
&
Performance remains strong on factors, HCF, multiples, LCM, and their combined use. Prime-number knowledge shifts from mixed understanding to correct classification in a basic concept-check context. Square-related knowledge is mixed: the student can evaluate a straightforward square calculation, but remains unstable on square-number properties and square-root/area interpretations.
&
Performance is mixed but shows a shift from uncertainty to partial understanding in square roots, cube roots, etc. The student succeeds on identifying square numbers and some basic square-number facts, but remains unstable when generalizing square-number properties. In number-factor work, the student shows secure understanding of factors, multiples, and common factors, but remains inconsistent on prime-number classification, especially the special case of 1.
\\

\textbf{Misconception}
&
\textcolor{CornellRed}{No recurring misconception is shown in the history.}
&
\textcolor{OliveGreen}{A misconception appears when the task requires generalizing from a familiar square-number fact to a rule about squares}: the student treats square-number properties as unstable or guesses a non-square result. 
\\

\textbf{Dialogue Acts}
&
\textit{Math Answer} is the dominant act, typically appearing as short guesses or revised answers after tutor prompting. The student also uses \textit{Acknowledge} fairly often, with occasional \textit{Not Understanding} and rare \textit{Off-Topic} comments.
&
\textit{Math Answer} is the dominant act, used for both tentative guesses and corrected responses. \textit{Acknowledge} appears as brief confirmations of tutor suggestions, \textcolor{OliveGreen}{\textit{Not Understanding} appears when the student is stuck, \textit{Off-Topic} appears occasionally as a brief social opener or distraction, and \textit{Seek Information} does not appear.}
\\

\textbf{Linguistic Style}
&
The student's tone is informal and subdued, sometimes slightly negative, but generally cooperative. Responses are very brief and conversational, with simple phrasing, spelling errors, and low-to-moderate confidence marked by hedging and confirmation-seeking.
&
The student uses an informal, conversational style with short, chat-like phrasing, casual spelling, and occasional texting-style abbreviations. Confidence is generally low to tentative, with uncertainty markers such as \textcolor{OliveGreen}{``i dont understand,'' ``ya,'' and ``am i correct,''} but becomes more direct when giving a revised answer.
\\
\bottomrule
\end{tabularx}
\caption{Head-to-head comparison between the GPT-generated profile and the GRPO-refined profile for the same student history. The GRPO-refined profile preserves the same KC-level knowledge state while providing more fine-grained evidence about misconceptions, dialogue acts, and linguistic style.}
\label{tab:profile_case_study}
\end{table*}

\section{Conclusions}

In this paper, we introduced the task of history-conditioned student simulation, and proposed a method for summarizing learning histories and faithfully simulating student dialogue turns, utilizing RL to refine each LLM. We demonstrate that our method effectively summarizes student learning histories, and outperforms baselines on student simulation across various metrics capturing student knowledge and behavior. We further show that profiles, and refining them with RL, are necessary for student simulation, and examine the effect of various aspects of profiles through ablations and qualitative analyses.

There are many avenues for future work. First, future work should study how to effectively include a student's entire learning history in context, potentially by leveraging interaction embeddings optimized for student simulation or using intelligent context retrieval. Second, future work should extend student simulation beyond dialogue turns, such as by predicting selected options or when a student requests help from a tutor. Third, future work should examine the faithfulness of simulations in fully simulated dialogues. Finally, future work should investigate this task in other domains, including programming and language learning.

\section*{Acknowledgments} This work is partially supported by Renaissance Philanthropy under the learning engineering virtual institute (LEVI) initiative and the NSF under grants 2153481, 2237676, and 2604250. 

\newpage
\section*{Limitations}

There are several limitations to our work. First, we only evaluate our task and methodology on a single private dataset in the math domain. While we are aware of one public dataset that may be applicable for our task, which we plan to investigate in the near future, there is a substantial lack of publicly available data containing student learning histories and dialogues, limiting the ability to test the generalizability of our method and task on other domains. Second, our method relies on a heuristic for trimming learning histories, only taking the most recent QA and dialogues, due to memory limitations in training. While this heuristic is effective, it is likely that more advanced techniques for selecting relevant interactions or extending the context length can improve performance, which we leave for future work. Finally, we only evaluate on turn-level student simulation in dialogues, following prior work. However, it will be important to examine how incorporating learning histories affects the faithfulness of fully simulated dialogues, although such evaluations will require new metrics to compare dialogues with simulated tutors and students to ground-truth dialogues.

\section*{Ethical Considerations}

There are several potential societal benefits and risks associated with our work. Faithful student simulation is a critical component of evaluating and training LLM tutors, and can enable more effective personalized tutoring for students. By improving the quality of LLM tutoring, it will be possible to improve the learning outcomes of students who do not have access to quality education or who require supplemental learning outside of the classroom. However, ubiquitous LLM tutoring runs the risk of replacing real teachers due to lower costs, which could not only replace jobs but harm student outcomes by removing critical human components from learning. We emphasize that we intend for LLM tutors built as a result of our work to be used only to supplement student learning experiences, rather than replace them entirely. Additionally, there is a risk of harm from bias due to inherent biases in data and pre-trained LLMs, where students from underrepresented populations could receive misaligned tutoring as a result of underrepresentation in the simulated student models. We strongly encourage any production systems to ensure that diverse demographics are included in training data, and that risks of bias are extensively evaluated before deploying with real students.

% \section*{Acknowledgments}

% Bibliography entries for the entire Anthology, followed by custom entries
%\bibliography{anthology,custom}
% Custom bibliography entries only
\bibliography{custom}

\appendix

\section{Related Work}
\label{sec:related-work}

\subsection{Profile-Conditioned Student Simulation}
Recent studies have explored student simulation in tutoring systems by conditioning LLMs on predefined student personas, learner profiles, or cognitive states \cite{wang-etal-2024-book2dial, mathdial, markel2023gpteach, liu-etal-2024-personality, litype, pan2025tutorup}. For example, \citet{liu2024socraticlm} defined five types of student cognitive states to encourage LLMs to simulate diverse student behaviors throughout the teaching process. Similarly, \citet{wang-etal-2025-training} leveraged an LLM-based student simulator to generate coding responses that reflect low, medium, and high levels of student knowledge. Unlike prior work that relies on predefined learner personas or profiles, \citet{Dong_Dai_Lv_Chen_2026} grounded student simulation in real learning trajectories from multi-round code submissions. However, their teacher-student interactions are still fully simulated from these trajectories, whereas our work learns student profiles directly from authentic educational data, using observed question-answering histories and real tutoring-dialogue histories to capture students' knowledge states and behavioral tendencies for next-turn simulation.

\subsection{Training Simulated Student Models}
Beyond prompting-based simulation, recent work has also explored SFT and preference optimization to improve the realism of simulated student responses. For instance, \citet{perczel2025teachlm} developed an authentic student model by applying parameter-efficient fine-tuning to LLMs using large-scale real student--tutor interaction data. \citet{cao2026developingauthenticsimulatedlearners} further compared multi-agent refinement, fine-tuning, and DPO for developing authentic simulated learners in mathematics teacher learning. Similarly, \citet{scarlatos2026simulatedstudentstutoringdialogues} systematically evaluated prompting, SFT, and DPO-based student simulation methods against real tutoring dialogues, highlighting the importance of aligning simulated students with real student behaviors. In addition, \citet{duan-etal-2026-kaser} investigated student error simulation to evaluate whether LLMs can generate errors that are consistent with students' underlying knowledge states.

Beyond student simulation, profile generation has also been studied in knowledge tracing. \citet{duan-etal-2026-automated} builds student profile on knowledge component mastery level. CIKT constructs a profile generator that summarizes students' question-answering histories into student profiles for knowledge tracing, and iteratively refines the generator based on reinforcement-style feedback from the predictor's correctness prediction performance \cite{li-etal-2025-cikt}. In contrast, our work focuses on history-conditioned student simulation rather than correctness prediction. We use both prior question-answering and tutoring-dialogue histories to learn compact profiles that capture students' knowledge states and individual behavioral tendencies over time. Additionally, we optimize the profile generator for next-turn simulation with GRPO and further train the simulator with DPO to improve profile-aligned generation.

\section{Additional Dataset Details}
\subsection{Dataset Details}
\label{apdx:ds_details}
We provide additional details on the dialogue dataset. The data is fully anonymized, so demographic information about individual students and tutors is not available.

Dialogues contain $12.95$ turns on average, counting both student and tutor turns. They can be initiated by either tutors or students, with $84.8\%$ initiated by tutors. Student and tutor turns contain $4.35$ and $13.15$ words on average, respectively, and include a mixture of English, numbers, mathematical symbols, and emojis. Each dialogue is associated with a set of ``subjects,'' which we use to define the knowledge components (KCs). The dataset contains interactions with $31$ tutors.

Following prior work~\cite{scarlatos2026simulatedstudentstutoringdialogues}, we use GPT-4.1~\cite{openai2gpt41} to annotate turn-level dialogue acts and correctness labels. The five dialogue-act labels are Math Answer, Not Understanding, Seek Information, Off-Topic, and Acknowledge. Across all valid act labels, there are $4,211$ Math Answer labels, $1,224$ Not Understanding labels, $1,012$ Seek Information labels, $1,657$ Off-Topic labels, and $2,524$ Acknowledge labels. For correctness annotation, we label student turns that attempt to answer a tutor-posed question as correct or incorrect, and assign NA to non-answer turns such as greetings and acknowledgments. Across the dataset, there are $2,811$ correct labels, $2,087$ incorrect labels, and $5,730$ NA labels.

\subsection{Potential Dataset for History-conditioned Student Simulation}
\label{apdx:assistments}
We additionally discuss a potential dataset as a promising candidate for future evaluation. The ASSISTments tutoring chat log dataset~\cite{prihar2023comparing} is another promising resource for history-conditioned student simulation. It contains tutoring request logs, message-level chat logs between students and tutors, problem-level dependent measures, sequence and problem metadata, and student action logs from ASSISTments. These components align closely with the requirements of our task: the chat logs provide authentic tutoring dialogues, the problem metadata and skill annotations provide KC information, and the dependent measures and action logs make it possible to construct prior question-answering and clickstream histories for each student. As a result, the dataset can support evaluation of whether our profile generator can summarize students' prior learning behavior and whether the simulator can use these profiles to predict future student turns in tutoring dialogues. We are currently running experiments on this dataset to test the generalizability of our method beyond the private math-learning-platform data used in the main experiments.

\begin{table*}[t]
\centering
\small
\setlength{\tabcolsep}{5pt}
\begin{tabular}{lccccc}
\toprule
Model & Acts & Corr. & Errors & Cos. Sim. & ROUGE-L \\
\midrule
SFT & 0.561 & 0.443 & 0.108 & 0.671 & 0.231 \\
History SFT & 0.584 & 0.472 & 0.116 & 0.681 & 0.235\\
DPO & 0.601 & 0.476 & 0.119 & 0.684 & 0.233 \\
ProfileRL (Ours) & 0.608 & 0.481 & 0.127 & 0.689 & 0.238 \\
\bottomrule
\end{tabular}
\caption{Section-level profile ablation results. The first row reports the full ProfileRL model, and each remaining row reports performance when the corresponding profile section is removed.}
\label{tab:smaller_model}
\end{table*}

\section{Additional Experimental Details}
\label{apdx:exp_details}
We provide additional experimental details for reproducibility. We use Llama-3.1-8B-Instruct~\cite{grattafiori2024llama3herdmodels} as the base model for all fine-tuned profile generators and student simulators. We use GPT-5.4~\cite{openai2026gpt54} for the prompting-based baselines and for initial profile generation.

We conduct preliminary hyperparameter exploration on the validation set, optimizing for our student-simulation metrics. We train the student simulator with AdamW, using a learning rate of $5\cdot 10^{-5}$, linear warmup for the first 10\% of training steps, an effective batch size of $64$ via gradient accumulation, weight decay of $1\cdot 10^{-2}$, and gradient norm clipping at $1.0$. We set the LoRA rank to $r=32$, $\alpha=64$, and dropout to $0.05$. For DPO on the student simulator, we use a learning rate of $5\cdot 10^{-6}$, $\beta=0.1$, $n=4$ candidate student utterances per turn, a mastery-level difference threshold of $\delta=0.4$. We compute the reward for each candidate turn as the average over the DPO scoring metrics described in Section~\ref{sec:student-simulator} and set a reward threshold of $\tau=0.5$. For overgeneration, we use a temperature of $1.0$ and top\_p of $0.95$. To avoid memory issues, we exclude prompts longer than $7,000$ characters during student simulator training. We train for $3$ epochs for SFT and $1$ epoch for DPO, and keep the student simulator frozen during RL-based profile generator refinement.

For the profile generator, we first train with SFT using the same optimizer settings as the student simulator: a learning rate of $5\cdot 10^{-5}$, linear warmup for 10\% of training steps, an effective batch size of $64$ via gradient accumulation, weight decay of $1\cdot 10^{-2}$, gradient norm clipping at $1.0$, LoRA rank $r=32$, $\alpha=64$, and dropout $0.05$. We then refine the profile generator with GRPO using a learning rate of $1\cdot 10^{-6}$, $\beta=0.1$, and a group size of $4$. We use vLLM~\cite{kwon2023efficientmemorymanagementlarge} with greedy decoding for both student-turn generation and profile generation, and set the maximum generation length to $500$ tokens. Student simulator training takes approximately $20$ minutes per SFT epoch and $60$ minutes per DPO epoch. Profile generator training takes approximately $30$ minutes for one SFT epoch and $90$ minutes for one GRPO epoch.

We also train an act classifier for the Acts metric, following the setup of prior work~\cite{scarlatos2026simulatedstudentstutoringdialogues}. The classifier reaches $92.4\%$ accuracy, indicating that it can reliably identify dialogue acts. We use GPT-5 mini~\cite{openai2025gpt5} as the judge model for Corr.\ and Errors; prior human evaluation shows that this model is reliable for these judgments~\cite{scarlatos2026simulatedstudentstutoringdialogues}. Finally, we use Qwen3-Embedding-8B~\cite{zhang2025qwen3embeddingadvancingtext} to compute cosine similarity.

\begin{table*}[t]
\centering
\small
\setlength{\tabcolsep}{5pt}
\begin{tabular}{lccccc}
\toprule
Profile Variant & Acts & Corr. & Errors & Cos. Sim. & ROUGE-L \\
\midrule
ProfileRL (Ours) & 0.605 & 0.466 & 0.150 & 0.689 & 0.271 \\
w/o Knowledge State & 0.587 & 0.411 & 0.104 & 0.673 & 0.260\\
w/o Knowledge Acquisition & 0.576 & 0.406 & 0.123 & 0.665 & 0.248 \\
w/o Misconception & 0.567 & 0.399 & 0.096 & 0.670 & 0.253 \\
w/o Dialogue Acts & 0.562 & 0.443 & 0.132 & 0.663 & 0.245 \\
w/o Linguistic Style & 0.566 & 0.440 & 0.134 & 0.662 & 0.243 \\
\bottomrule
\end{tabular}
\caption{Section-level profile ablation results. The first row reports the full ProfileRL model, and each remaining row reports performance when the corresponding profile section is removed.}
\label{tab:profile-section-ablation}
\end{table*}

We report ROUGE-L using the Hugging Face evaluate implementation of ROUGE. To the best of our knowledge, the software and models used in our implementation either have open-source licenses or do not list a license. In addition, our use of these resources, including the OpenAI API, is consistent with their intended terms of use. We use GPT-4.1 for data annotation, GPT-5 mini for Corr. and Errors. evaluation, and GPT-5.4 for all other purposes. If we release code, we will ensure that the license and terms reflect the sources on which our implementation builds. We use ChatGPT to assist with writing and editing the manuscript.

\section{Additional Quantitative Evaluation}
\subsection{Generalization to a Smaller Model}
\label{appdix: smaller_model}

We conduct additional experiments using a smaller student simulator, Llama-3.2-3B, to further assess the generalizability of ProfileRL. We report the results in Table~\ref{tab:smaller_model}. Overall, ProfileRL achieves the best performance across all metrics, outperforming SFT, History SFT, and DPO and all prompting-based baselines. These results further demonstrate that our proposed method can generalize to a smaller, more lightweight model.

\subsection{Additional Ablation Study}
\label{appdix: add_ablation}

\begin{table}
\centering
\small
\begin{tabular}{lc}
\toprule
History Window Setting & Target-KC Coverage \\
\midrule
$N^{\mathrm{QA}}=4$  & 44.9\% \\
$N^{\mathrm{QA}}=8$  & 47.3\% \\
$N^{\mathrm{QA}}=16$ & 49.4\% \\
$N^{\mathrm{QA}}=32$ & 53.7\% \\
$N^{\mathrm{QA}}=\mathrm{all}$ & 63.8\% \\
\midrule
$N^{\mathrm{D}}=1$ & 19.3\% \\
$N^{\mathrm{D}}=2$ & 20.6\% \\
$N^{\mathrm{D}}=3$ & 21.6\% \\
$N^{\mathrm{D}}=4$ & 21.8\% \\
$N^{\mathrm{D}}=5$ & 22.0\% \\
$N^{\mathrm{D}}=\mathrm{all}$ & 22.6\% \\
\bottomrule
\end{tabular}
\caption{
Target-KC coverage under different history window settings.
}
\label{tab:history-window-coverage}
\end{table}

Table~\ref{tab:profile-section-ablation} examines the contribution of each profile section. Removing any single section lowers performance compared with the full ProfileRL model, suggesting that faithful simulation benefits from combining knowledge state, knowledge acquisition, misconception, dialogue-act, and linguistic-style information. Among the knowledge-related sections, removing \textit{Misconception} produces the largest drops in Corr.\ and Errors, indicating that explicit misconception information is especially important for reproducing whether students answer correctly and whether they make the same underlying mistake. Among the behavior-related sections, removing \textit{Dialogue Acts} yields the lowest Acts score, while removing \textit{Linguistic Style} produces the largest drops in Cos.\ Sim.\ and ROUGE-L.

\begin{figure*}[t]
    \centering
    \begin{subfigure}{0.48\linewidth}
        \centering
        \includegraphics[width=\linewidth]{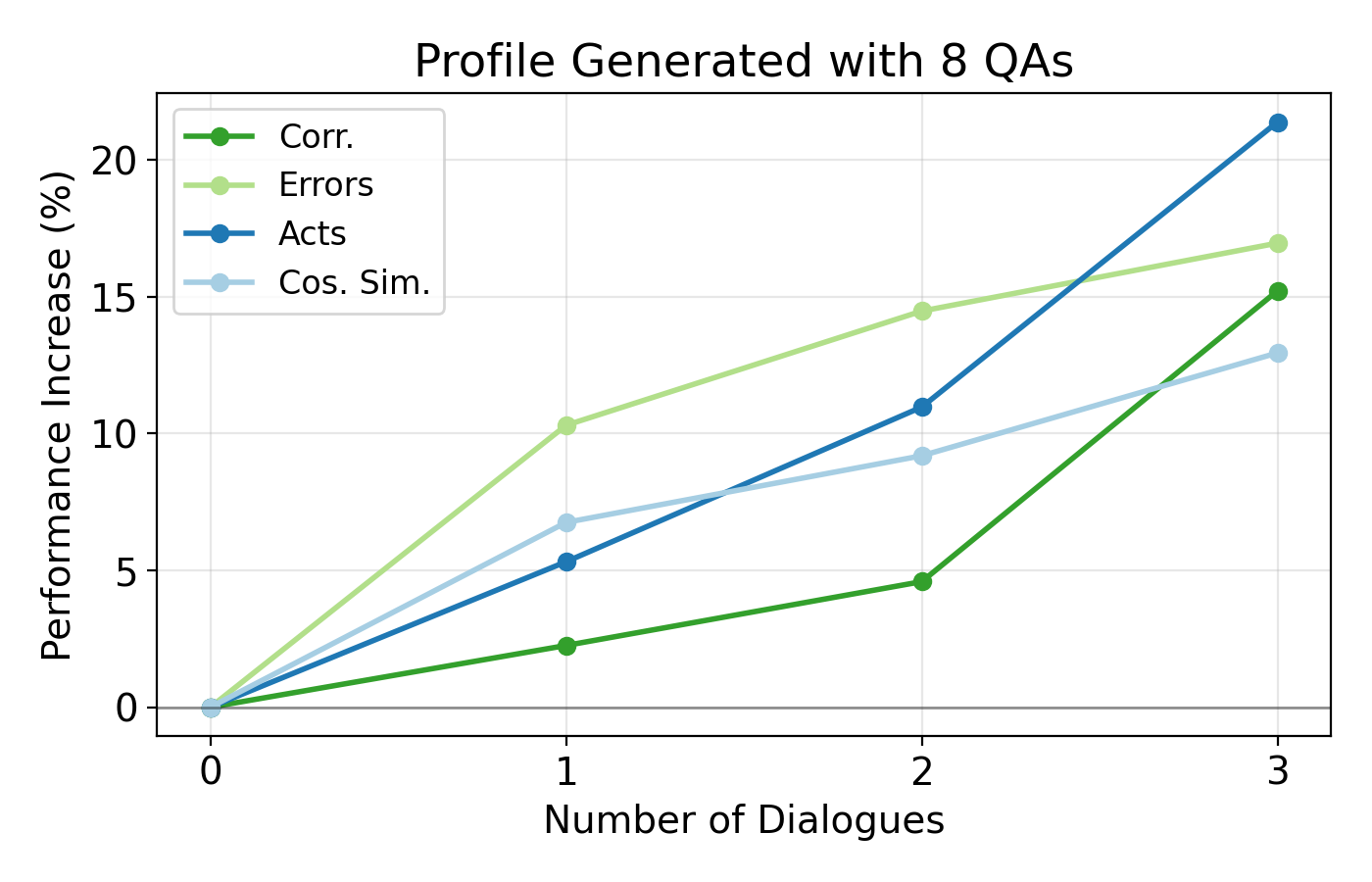}
        % \caption{Dialogue context ablation}
        \label{fig:context-ablation-dialogues}
    \end{subfigure}
    \hfill
    \begin{subfigure}{0.48\linewidth}
        \centering
        \includegraphics[width=\linewidth]{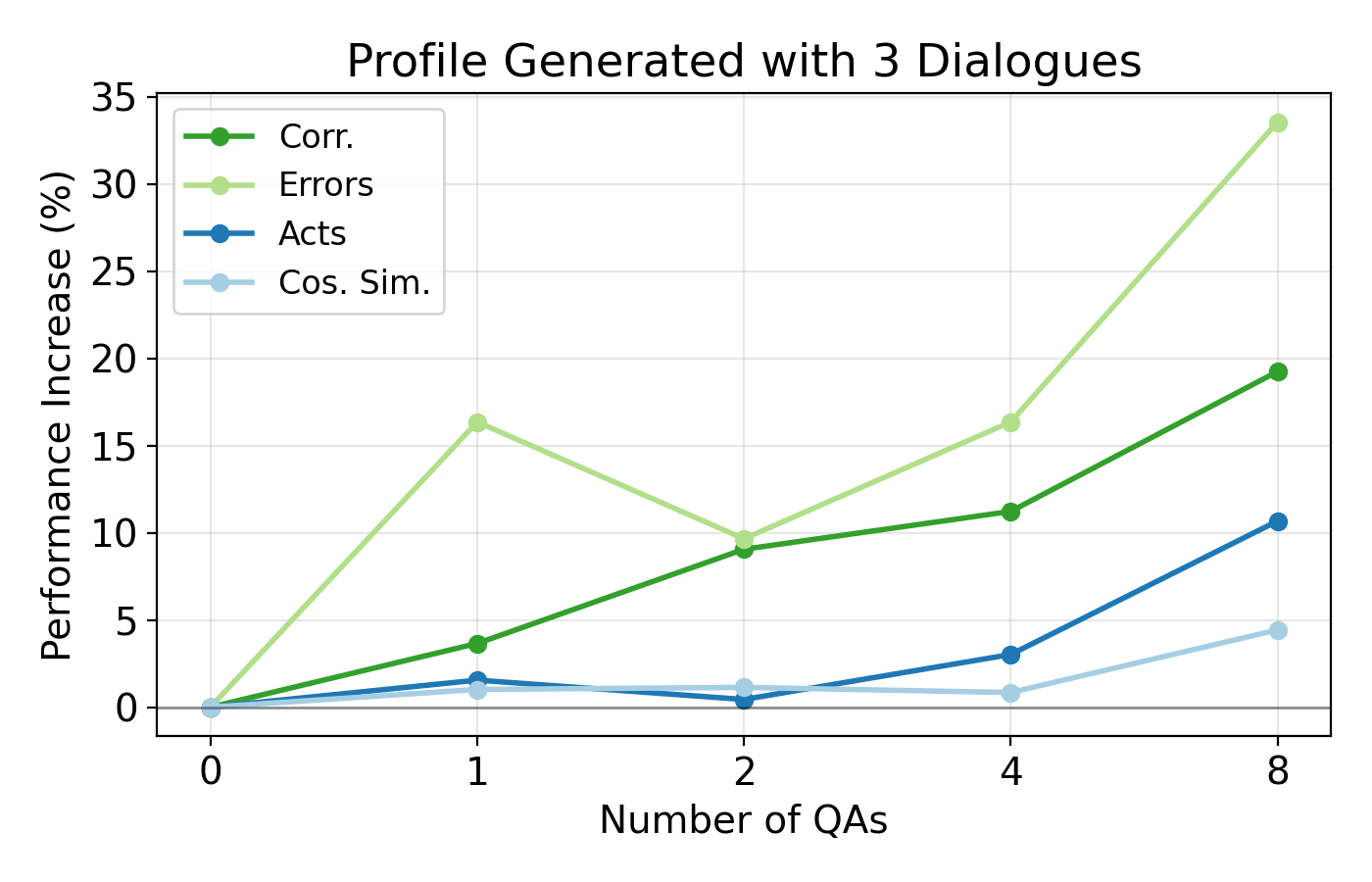}
        % \caption{QA context ablation}
        \label{fig:context-ablation-qas}
    \end{subfigure}
    \vspace{-0.8cm}
    \caption{Context ablation for profile generation. We vary the amount of historical question-answering records and dialogue context used to construct student profiles and evaluate the resulting student simulation performance across Acts, Correctness, Errors, and Cos.\ Sim. The left subfigure fixes the number of question-answering records to 8 and varies the number of dialogues, while the right subfigure fixes the number of dialogues to 3 and varies the number of question-answering records}
    \label{fig:context-ablation}
\end{figure*}

We next study how the amount of historical context used for profile generation affects downstream student simulation performance. Specifically, we vary the number of question-answering records and prior dialogues used to construct the profile. Results are shown in Figure~\ref{fig:context-ablation}.

Overall, the strongest performance is achieved by our main setting, which uses $8$ question-answering records and $3$ dialogues for profile generation. Increasing the number of dialogues consistently improves Acts, suggesting that additional dialogue history provides a more complete representation of students' dialogue-act tendencies. Cos.\ Sim.\ also improves as more dialogues are included, indicating that dialogue history helps capture student-specific linguistic patterns. The knowledge-related metrics, Corr.\ and Errors, also improve with more dialogue history, suggesting that prior tutor-student interactions provide additional evidence about students' understanding and misconceptions.

When the number of dialogues is fixed, the number of historical question-answering records has a stronger effect on Corr.\ and Errors. As more question-answering records are included, both metrics improve, indicating that richer question-answering history helps the profile better represent students' knowledge states and misconception patterns. At the same time, Acts also improves, suggesting that knowledge history can help predict how students respond in dialogue. Together, these ablations again show that the two sources of historical context are complementary: dialogue history contributes to knowledge-related metrics, while question-answering history also benefits behavior-related metrics. This finding further supports the design of our profile-based student simulation method, which combines both types of historical information to encode student knowledge and behavior.

\section{Qualitative Case Studies}
\label{apdx:add_qualitative}

Tables~\ref{tab:profile_case_study} and~\ref{tab:add_profile_case_study} present two head-to-head comparisons between GPT-generated profiles and GRPO-refined profiles. Within each table, the two profiles are generated from the same student history. Across both examples, GRPO refinement turns broad profile summaries into more diagnostic representations for downstream simulation. In particular, the GPT-generated profiles miss recurring misconception patterns even when the history contains incorrect question-answering records, whereas the GRPO-refined profiles identify KC-specific weaknesses and describe the corresponding misconception. This information is important for simulating whether a student answers correctly, whether they reproduce the same error, and how they respond in dialogue. In addition, the GRPO-refined profiles include more concrete linguistic evidence, such as observed phrases and stylistic markers from the student's prior dialogue turns, giving the simulator clearer cues about word choice, brevity, and uncertainty.

\begin{table*}[t]
\centering
\small
\setlength{\tabcolsep}{4pt}
\renewcommand{\arraystretch}{1.15}
\begin{tabularx}{\textwidth}{p{0.15\textwidth} X X}
\toprule
\textbf{Profile Section} & \textbf{GPT Profile} & \textbf{GRPO-refined Profile} \\
\midrule
\textbf{Knowledge State}
&
\textcolor{CornellRed}{2D Names and Properties of Shapes-Others: 0/2}; Averages (mean, median, mode) from a List of Data: 1/1; Place Value: 1/2; Angles in Triangles: 1/1; Properties of Quadrilaterals: 1/1; Adding and Subtracting Negative Numbers: 1/1.
&
\textcolor{OliveGreen}{2D Names and Properties of Shapes-Others: 0/2}; Averages (mean, median, mode) from a List of Data: 1/1; Place Value: 1/2; Angles in Triangles: 1/1; Properties of Quadrilaterals: 1/1; Adding and Subtracting Negative Numbers: 1/1.
\\

\textbf{Knowledge Acquisition}
&
Averages (mean, median, mode) from a List of Data strengthened to consistent success, Place Value shifted from mixed performance to success, and 2D Names and Properties of Shapes-Others remained weak across comparable regular-polygon judgments. Angles in Triangles, Properties of Quadrilaterals, and Adding and Subtracting Negative Numbers held steady at correct performance, while no demonstrated history is available for Linear Equations or Multiplying Terms beyond help-seeking in dialogue.
&
Averages (mean, median, mode) from a List of Data shows partial but improving understanding: the student can identify the middle value in a small set when prompted, but performance remains unstable when the task requires explaining the median concept. Place Value shows a shift from confusion with place-value multiplication to correct handling of multiplying by 100, while Angles in Triangles and Properties of Quadrilaterals appear secure from the available evidence. Adding and Subtracting Negative Numbers is secure, and 2D Names and Properties of Shapes-Others remains weak with no demonstrated improvement.
\\

\textbf{Misconception}
&
\textcolor{CornellRed}{No recurring misconception is shown in the history.}
&
\textcolor{OliveGreen}{A recurring misconception appears when the question requires explaining a shape property or mathematical concept}: the student treats the task as selecting a label or recalling a rule rather than justifying the underlying idea, producing short answer fragments that lack explanation. 
\\

\textbf{Dialogue Acts}
&
Math Answer is the dominant act in tutoring, used both to give direct answers and to ask answer-checking questions, while Acknowledge is also frequent and typically appears as brief confirmations that close or move along the exchange. There is essentially no Not Understanding or Seek Information labeling, with help requests instead phrased through short answer-oriented math utterances.
&
Math Answer is the dominant act, typically used both for direct answers and for asking for answer choices, while Acknowledge is also frequent and used to close the exchange or confirm understanding. Off-Topic appears occasionally as brief social or greeting-like turns, and Not Understanding and Seek Information is absent; the student does not usually ask for clarification or express confusion.
\\

\textbf{Linguistic Style}
&
The student’s tone is casual and polite, with brief social niceties and an informal, conversational manner. Their responses are very short and minimally elaborated, using simple lowercase phrasing and showing moderate confidence through quick confirmations and answer-checking rather than detailed explanation.
&
The student uses a casual, informal style with short, lower-case messages, texting-style spellings, and conversational phrasing. Confidence is variable: they can sound tentative when explaining reasoning shown through \textcolor{OliveGreen}{hedging phrases like “maybe”}, but more direct and decisive when giving a quick answer or closing the exchange.
\\
\bottomrule
\end{tabularx}
\caption{Additional Head-to-head comparison between the GPT-generated profile and the GRPO-refined profile.}
\label{tab:add_profile_case_study}
\end{table*}

\section{History Window Size Analysis}
\label{app:history-window}
To choose suitable values for $N^{\mathrm{QA}}$ and $N^{\mathrm{D}}$, we measure target-KC coverage under different history window sizes. This analysis captures the trade-off between retaining relevant historical evidence and keeping the context length tractable. We define target-KC coverage as the proportion of student-turn prediction examples whose selected history contains at least one prior interaction associated with the KC of the current tutor-posed task to which the student simulator responds. Higher coverage indicates that the selected history is more likely to contain information relevant to simulating the student's next response, while larger windows also increase prompt length and computational cost.

As shown in Table~\ref{tab:history-window-coverage}, increasing the QA window from $N^{\mathrm{QA}}=4$ to $N^{\mathrm{QA}}=8$ improves target-KC coverage from 44.9\% to 47.3\%. Further increasing the QA window yields additional coverage, reaching 49.4\% for $N^{\mathrm{QA}}=16$ and 53.7\% for $N^{\mathrm{QA}}=32$, but with substantially higher context and computational costs. In our implementation, setting $N^{\mathrm{QA}}=16$ exceeded the available memory budget during model training. For dialogue interactions, increasing the window from $N^{\mathrm{D}}=1$ to $N^{\mathrm{D}}=3$ improves target-KC coverage from 19.3\% to 21.6\%. Larger dialogue windows provide only marginal additional coverage, reaching 21.8\% for $N^{\mathrm{D}}=4$, 22.0\% for $N^{\mathrm{D}}=5$, and 22.6\% when using all available dialogue history. Based on this analysis, we set $N^{\mathrm{QA}}=8$ and $N^{\mathrm{D}}=3$ as a practical balance between retaining relevant historical information and keeping profile generation tractable.

\section{Prompts}
\label{sec:prompts}
\begin{figure*}[!tbhp]
    \centering
    \includegraphics[width=1\linewidth]{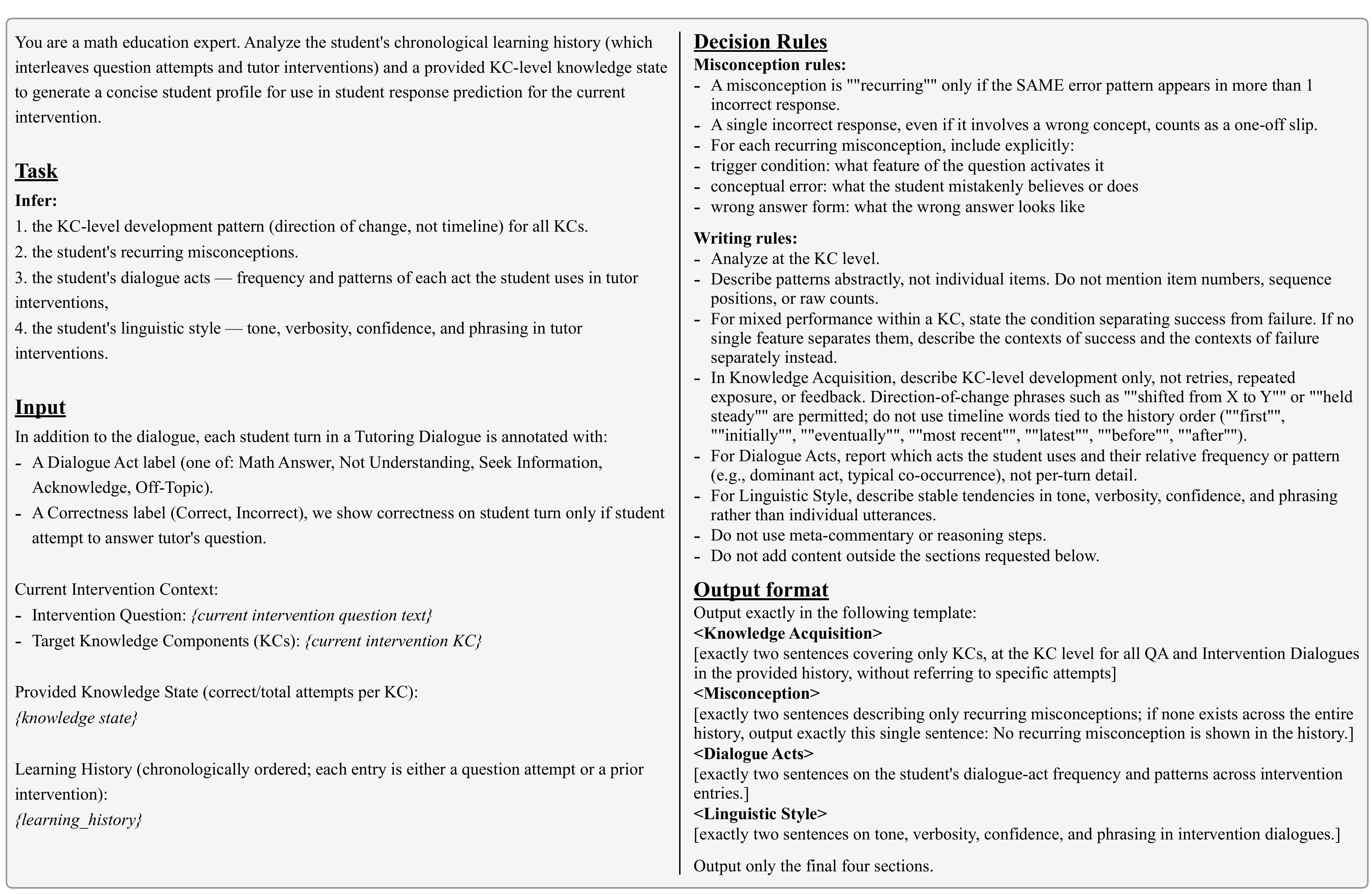}
    \caption{Prompt for initial profile generation.}
    \label{fig:profile_prompt}
\end{figure*}

We show the prompt we use for initial prompt generation in Figure~\ref{fig:profile_prompt}.

\end{document}